\documentclass[journal]{IEEEtran}
\usepackage{cite}
\usepackage{amsmath,amssymb,amsfonts}
\usepackage{algorithmic}
\usepackage{color}
\usepackage{makecell}
\usepackage{graphicx}
\usepackage{textcomp}
\usepackage{diagbox}
\usepackage{amsmath}
\usepackage{autobreak}
\usepackage{bbding}
\usepackage{multirow}
\usepackage{hyperref}

\definecolor{ref}{rgb}{0,0.541,0.855}

\begin{document}
\title{Dispensed Transformer Network for Unsupervised Domain Adaptation}
\author{Yunxiang Li, Jingxiong Li, Ruilong Dan, Shuai Wang, Kai Jin, Guodong Zeng, Jun Wang, Xiangji Pan, Qianni Zhang, Huiyu Zhou, Qun Jin, \IEEEmembership{Senior Member, IEEE}, Li Wang, \IEEEmembership{Senior Member, IEEE}, Yaqi Wang
\thanks{This work was supported by the National Key Research and Development Program of China (Grant No. 2019YFC0118404). (Yunxiang Li and Jingxiong Li are co-first authors.) (Corresponding author: Yaqi Wang.) }
\thanks{Yunxiang Li, Ruilong Dan are with College of Computer Science and Technology, Hangzhou Dianzi University, Hangzhou, China (e-mail: li1124325213@hdu.edu.cn, drldyx20xx@hdu.edu.cn).}
\thanks{Jingxiong Li is with Artificial Intelligence and Biomedical Image Analysis Lab, Westlake University, Hangzhou, China (e-mail: lijingxiong@westlake.edu.cn).}
\thanks{Shuai Wang is with School of Mechanical, Electrical and Information Engineering, Shandong University, Weihai, China (e-mail: shuaiwang@sdu.edu.cn).}
\thanks{Kai Jin, Xiangji Pan are with Department of Ophthalmology, the Second Affiliated Hospital of Zhejiang University, Hangzhou, China (e-mail: jinkai@zju.edu.cn, pan\_xiangji@zju.edu.cn).}
\thanks{Guodong Zeng is with sitem Center for Translational Medicine and Biomedical Entrepreneurship, University of Bern, Bern, Switzerland (e-mail: guodong.zeng@sitem.unibe.ch).}
\thanks{Jun Wang is with School of Biomedical Engineering, Shanghai Jiao Tong University, Shanghai, China (e-mail: wjcy19870122@sjtu.edu.cn).}
\thanks{Qianni Zhang is with School of Electronic Engineering and Computer Science, Queen Mary University of London, London, UK (e-mail: qianni.zhang@qmul.ac.uk).}
\thanks{Huiyu Zhou is with School of Computing and Mathematical Sciences, University of Leicester, UK (e-mail: hz143@leicester.ac.uk).}
\thanks{Qun Jin is with the Department of Human Informatics and Cognitive Sciences, Faculty of Human Sciences, Waseda University, Tokyo, Japan (e-mail: jin@waseda.jp).}
\thanks{Li Wang is with Department of Radiology and Biomedical Research Imaging Center, University of North Carolina at Chapel Hill, Chapel Hill, USA (e-mail: li\_wang@med.unc.edu).}
\thanks{Yaqi Wang is with the College of Media Engineering, Communication University of Zhejiang, Hangzhou, China (e-mail: wangyaqi@cuz.edu.cn).}}

\maketitle

\begin{abstract}
Accurate segmentation is a crucial step in medical image analysis and applying supervised machine learning to segment the organs or lesions has been substantiated effective. However, it is costly to perform data annotation that provides ground truth labels for training the supervised algorithms, and the high variance of data that comes from different domains tends to severely degrade system performance over cross-site or cross-modality datasets. To mitigate this problem, a novel unsupervised domain adaptation (UDA) method named dispensed Transformer network (DTNet) is introduced in this paper. Our novel DTNet contains three modules. First, a dispensed residual transformer block is designed, which realizes global attention by dispensed interleaving operation and deals with the excessive computational cost and GPU memory usage of the Transformer. 
Second, a multi-scale consistency regularization is proposed to alleviate the loss of details in the low-resolution output for better feature alignment.
Finally, a feature ranking discriminator is introduced to automatically assign different weights to domain-gap features to lessen the feature distribution distance, reducing the performance shift of two domains.
The proposed method is evaluated on large fluorescein angiography (FA) retinal nonperfusion (RNP) cross-site dataset with 676 images and a wide used cross-modality dataset from the MM-WHS challenge. Extensive results demonstrate that our proposed network achieves the best performance in comparison with several state-of-the-art techniques.

\end{abstract}

\begin{IEEEkeywords}
Transformers in Vision, Unsupervised Domain Adaptation, Segmentation
\end{IEEEkeywords}

\section{Introduction}
\label{sec:introduction}
\IEEEPARstart{M}{edical} imaging is mainly used for quantitative and accurate diagnosis, severity estimation, and treatment planning. Accurate segmentation of medical images is a significant step during radiotherapy treatment. Unfortunately, precise segmentation relies heavily on human experts to deal with easily confused visual features and scattered deformable lesions, and it is extremely time-consuming and error-prone. Therefore, an automatic and reliable segmentation method is necessary. Medical research has achieved remarkable advances in high-performance segmentation models based on deep learning. Notwithstanding the new performance peaks, they still require large and high-quality annotated datasets, particularly in medical imaging, where the annotations are expensive to acquire. 

For the above reasons, if the new domain data appears, supplying a large amount of annotation to each new domain manually is impractical, and it is preferred to conduct automated segmentation of new site or modality data with the existing annotations. However, there is a significant domain-gap between the data from different sites or modalities, leading to a large domain-shift performance loss. As domain shift largely undermines the performance of deep learning models\cite{tommasi2016learning}, it is a thriving field to design methods that are expected to learn and transfer robust knowledge between domains. Different from supervised learning, unsupervised domain adaptation (UDA) transfers knowledge from a labelled domain (source domain) to an unlabeled domain (target domain). To be specific, first of all, the model is trained with the supervised loss on the source domain. The discriminator is then adopted to reduce the difference between the source and target domains to be minimum whilst maintaining the consistency of feature distributions. 

\begin{figure}[ht]
\centering
  \includegraphics[width=.5\textwidth]{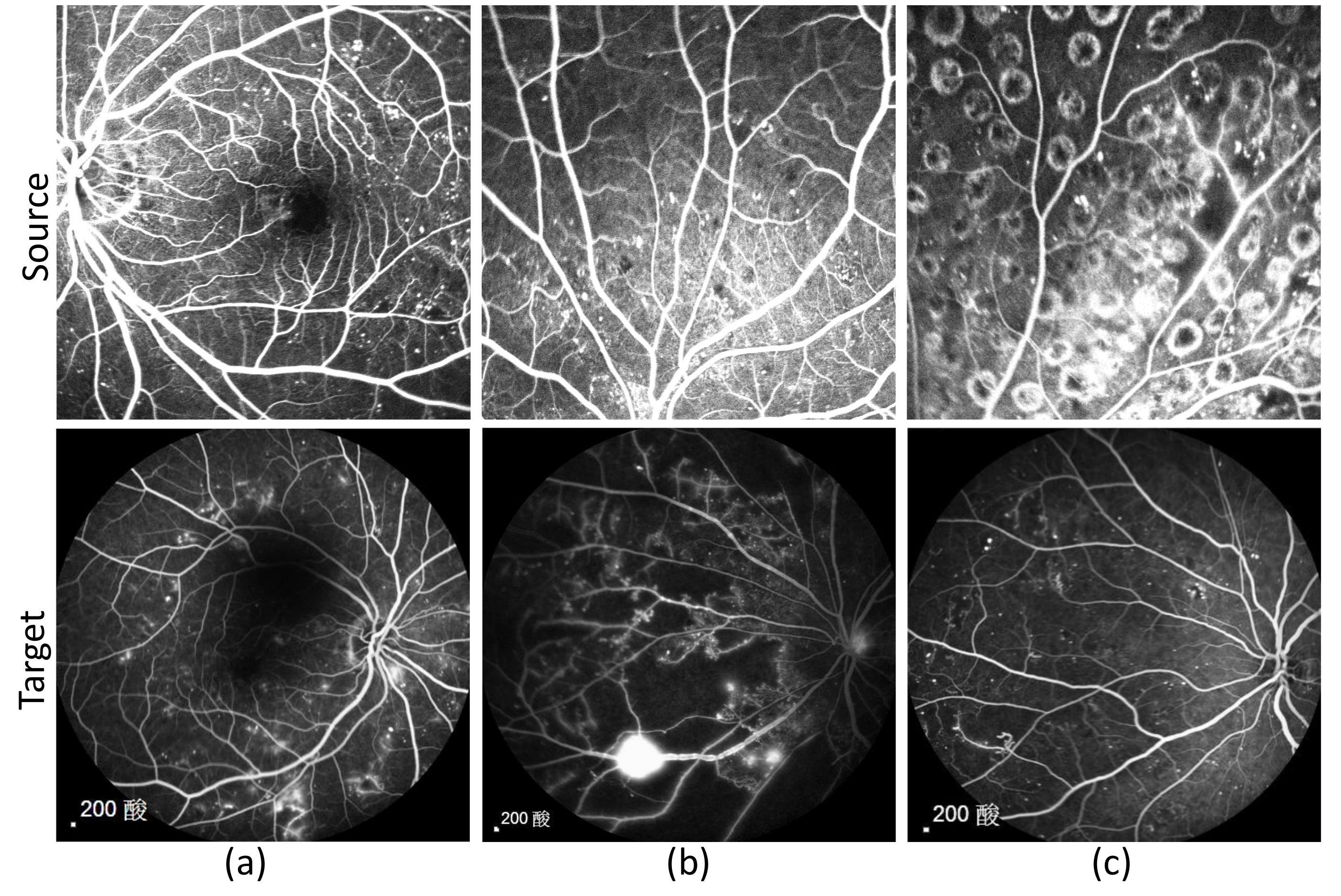}
  \caption{Cross-site fluorescein angiography retinal nonperfusion segmentation dataset. Images in the first and the second rows show some examples of the two sites. [Please indicate the data source]}
  \label{fig:challenge}
\end{figure}

Generally, UDA is mainly used for two different applications, i.e. cross-site and cross-modality learning. \textcolor{ref}{Fig. \ref{fig:challenge}} shows the distinctive cross-site dataset from two typical fluorescein angiography (FA) images collected by different devices and hospitals. As the retina is a curved surface, there are illumination and focusing inconsistencies between the periphery and the center of the image, thus FA images from different sites have different view fields, appearances, contrasts, and patterns. \textcolor{ref}{Fig. \ref{fig:challenge}} shows the performance shift of the two domains. Specifically, the source images with $30^\circ$ fields of view provide a detailed view of macular with unique features such as laser dots caused by retinal photocoagulation, shown in \textcolor{ref}{Fig. \ref{fig:challenge} Source (c)}. The target domain contains images photoed in the $55^\circ$ view field, generating a comprehensive circular view of the retina, consisting of an optic disc area with lower contrast and lightness. 

\begin{figure}[ht]
\centering
  \includegraphics[width=.5\textwidth]{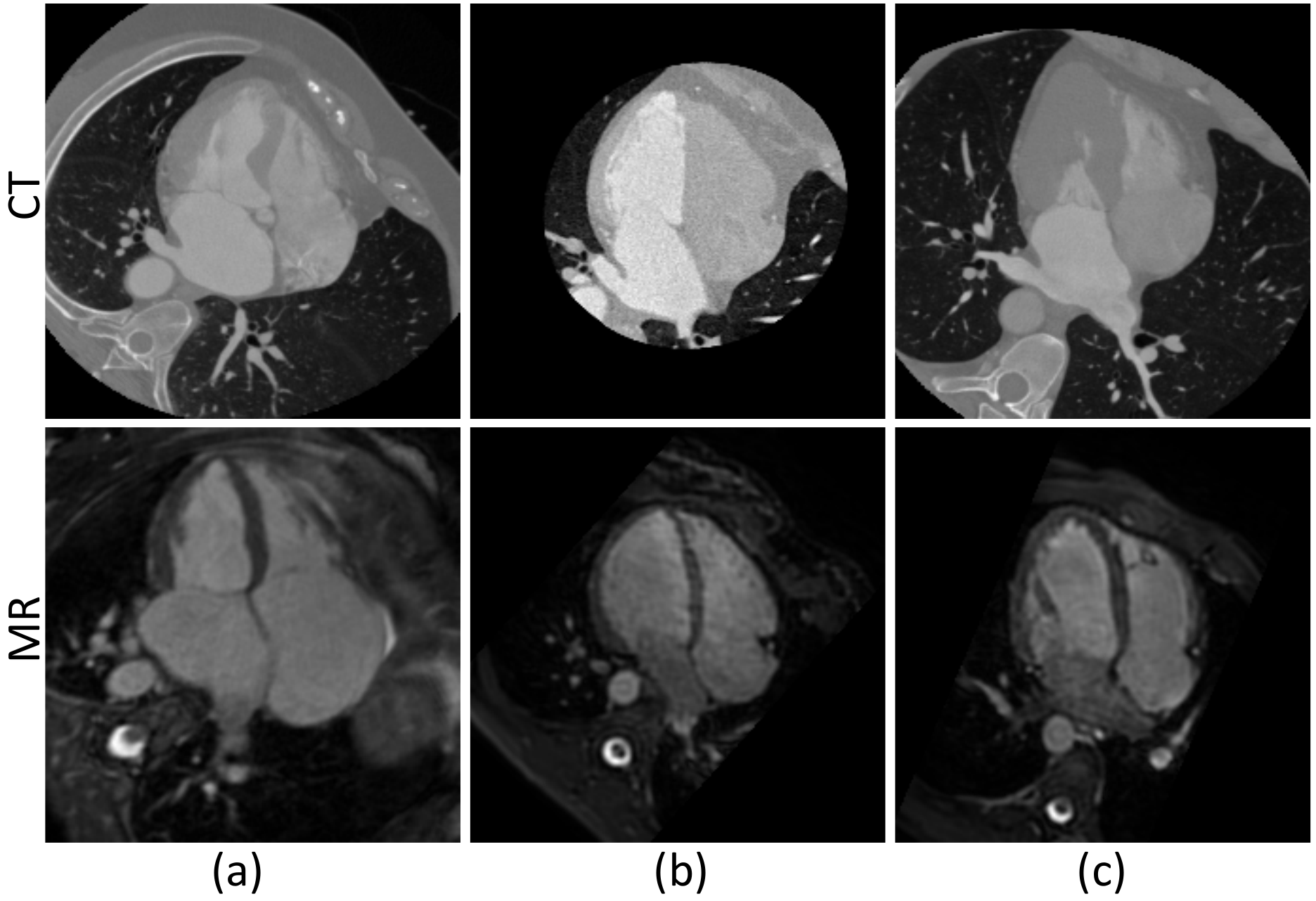}
  \caption{Cross-modality cardiac segmentation dataset. Images in the first and the second rows show some examples of two modalities, i.e. CT and MR. [Please indicate the data source]}
  \label{fig:ctmr}
\end{figure}

Another widely used UDA application is for cross-modality data, and \textcolor{ref}{Fig. \ref{fig:ctmr}} provides examples of the cropped 2D images of CT and MR in cardiac segmentation. 
The data of different modalities has a larger domain-gap, which is more challenging. It is also worth noting that the cardiac cross-modality segmentation data have similar shapes, while retinal nonperfusion in the FA images has no fixed shape, and it is difficult to design a network structure that performs well on both cross-site and cross-modality applications.

Nowadays, the widely used UDA methods are based on the convolutional neural network (CNN). Due to the inherent limitations of locality, CNN pays more attention to local features, such as local texture details, which causes  performance loss due to the domain shift. Transformer is a non-local operation completely different from convolution, which can realize the attention of global features. Nevertheless, it is difficult to apply Transformer directly to computer vision tasks due to the excessively computational cost and GPU memory usage. On the premise of implementing the global operation of Transformer, it is necessary to reduce the computational cost and memory usage as much as possible.

Following these requirements, an end-to-end dispensed Transformer network (DTNet) is constructed in this paper. 
The main contributions of this paper are listed as follows: 

\begin{itemize}
    \item Dispensed residual transformer block is introduced to achieve both global operation and computational efficiency for the Transformer.
    \item Multi-scale consistency regularization is proposed to ensure that the feature maps have optimal predictions and to alleviate the loss of details in the low-resolution outputs for better feature alignment in the source domain training.
    \item In the progress of feature alignment, our proposed feature ranking discriminator is implemented to reduce the domain gap. 
\end{itemize}

\section{Related works}
In this paper, we propose an unsupervised domain adaptation method namely dispensed Transformer network for retinal nonperfusion and cardiac segmentation. This research mainly involves three fields: medical image segmentation, unsupervised domain adaptation, and transformer in vision. Thus, the works related to these three fields are reviewed in this section. 

\subsection{Medical Image Segmentation}
Medical image segmentation plays a key role in computer-aided medicine due to the increasing demand for diagnostic precision and efficacy. Recently, deep learning-based methods have shown remarkable successes in diverse medical image analysis tasks. Ronneberger et al.\cite{ronneberger2015u} proposed a fully convolutional network called U-Net for biomedical image segmentation. 
To assist nonperfusion segmentation, Joan et al. \cite{rio2020deep} report a U-Net based algorithm to generate a quantitative evaluation for the Ultra-Wide-Field FA images. Tang et al. \cite{tang2021automated} proposed a feature extractor based on convolutional neural network (CNN) and then establish a model to learn from multiple labels. To segment cardiac structures, Isensee et al. \cite{isensee2017automatic} used an ensemble of U-Net inspired architectures. Schlemper et al. \ cite{schlemper2018cardiac} presented a framework with end-to-end synthesis and latent feature interpolation techniques for cardiac MR segmentation. Zreik et al.\cite{zreik2016automatic} proposed a two-step network for cardiac substructures segmentation. However, the majority of the segmentation methods operated on medical images highly rely on accurate annotations and low domain shift assumption, which is impractical with a large amount of data involved.

\begin{figure*}[ht]
\centering
  \includegraphics[width=1.\textwidth]{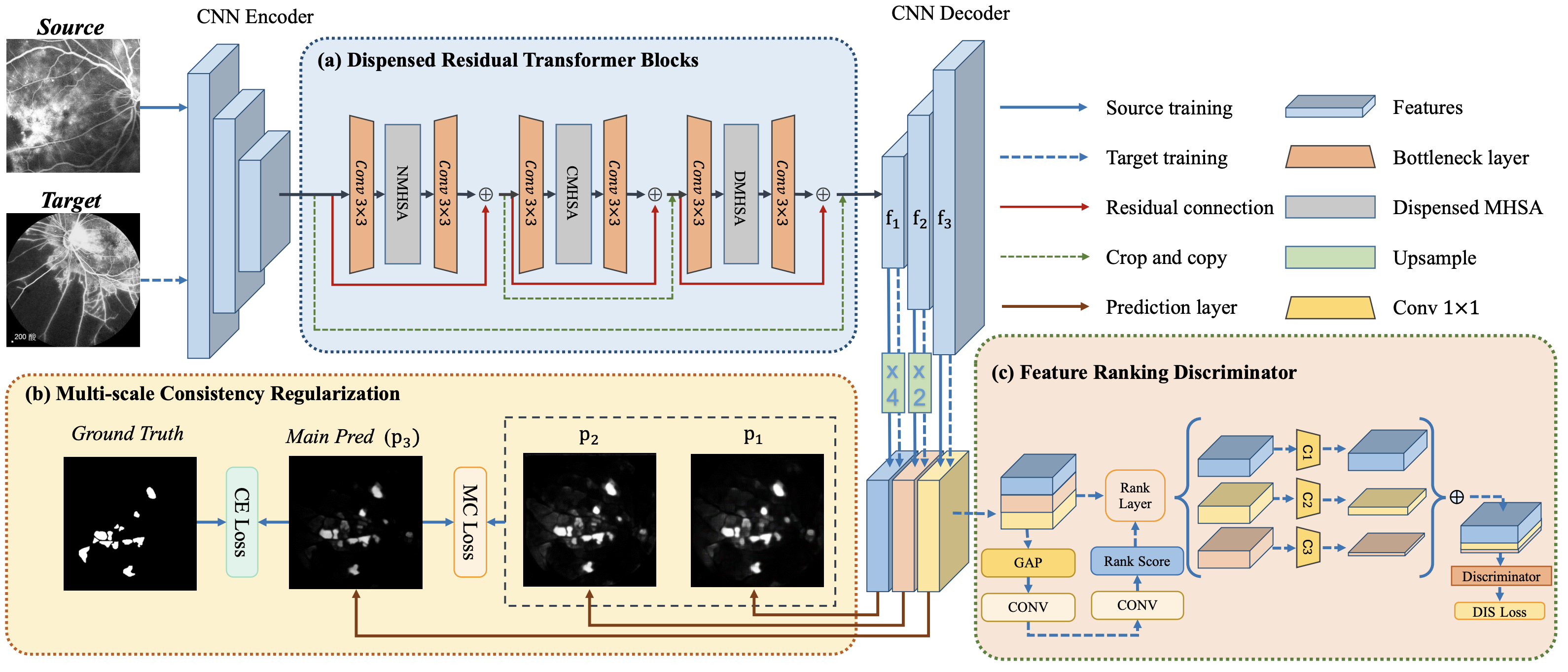}
  \caption{Overview of the proposed approach with multiscale consistency loss (MC loss) and cross entropy loss (CE loss). The max-pooling and up conv of the U-shaped structure are omitted in the figure for simplicity. }
  \label{fig:network}
\end{figure*}

\subsection{Unsupervised Domain Adaptation}

In medical image analysis, domain adaptation is a thriving research field aiming at perfectly transferring knowledge from the source domain to the target domain \cite{ghafoorian2017transfer}. Deep learning-based models have been proposed to reduce the performance loss caused by cross-modality, inter-scanner, or multi-center variations \cite{ren2021segmentation}. Unsupervised domain adaptation is a topic aiming at improving the performance of deep models in the target domain by using only source domain annotations. Various works are deployed, which can be divided into discrepancy-based and adversarial models \cite{zhao2020a}. 

Discrepancy-based approaches usually have similar network streams for source and target domains, and focus on reducing inter-domain distance by forcing the networks to extract domain invariant features from a large amount of image data. 
For example, Carlucci et al. \cite{carlucci2019domain} proposed a siamese network and minimized the discrepancy by aligning their batch normalization results. Wu et al. \cite{wu2020cf} introduced an explicit domain adaptation method named CFDnet to reduce the distribution discrepancy by a metric based on characteristic functions of distributions. However, these methods of minimizing the discrepancy between domains are still struggling when the domain gap is too large, severely limiting their broad applicability. Moreover, some discrepancy-based approaches demonstrate a weak capacity in backpropagation, especially for segmentation tasks with high latent space dimensions.

Adversarial models tackle the above problem by adopting the conceptions of generative adversarial networks (GANs) \cite{goodfellow2014generative}. One of the research studies is to transfer the images from the target domain to the source domain, and then to apply the source-trained model to the pre-processed images \cite{zhang2018task}. The key step of these approaches is to generate style-realistic images in another domain with unpaired data. CycleGAN \cite{zhu2017unpaired} is successful on image-to-image style translation. However, these methods ignore the prior that, for segmentation tasks, the label space of different domains is usually highly correlated w.r.t. the anatomical structures. Therefore, it is a better solution to deploy domain discriminators to enforce the cross-domain features to be aligned in terms of spatial structures and distribution \cite{dou2019pnp}. Jiang et al. \cite{jiang2020psigan} developed a joint probabilistic segmentation and distribution matching method namely PSIGAN, which has demonstrated the state-of-the-art performance in the task of cross-modality unsupervised domain adaptation. 
However, the above methods are based on convolutional neural networks, which have inherent limitations of locality. Convolution pays more attention to local features, such as local texture details, resulting in a loss of system performance in the domain shift.


\subsection{Transformers in Vision}
CNNs have the inherent drawback in modelling wide-ranging information, which is conflicting with the increasing demand of the models that are capable of capturing long-range features. 
Wang et al.\cite{wang2018non} introduced non-local operations for long-range dependencies. Cao et al.\cite{cao2019gcnet} improved their work and presented Global Context Network (GCNet), which is lightweight and capable of modeling global features. However, these methods are still based on convolution.
Transformer \cite{vaswani2017attention} is proposed for natural language processing tasks at first, providing CNN free structures that can achieve wide-ranging dependencies, where multi-head self-attention (MHSA) has excellent dynamic weighting properties. Its efficiency in modeling wide-ranging contextual information has recently achieved great successes in various vision tasks \cite{yuan2021ocnet,cao2021swin}. 
Alexey et al. \cite{dosovitskiy2020image} proposed a novel network called Vision Transformer (ViT) that achieves decent performance compared with CNNs on large image datasets. ViT treats the input image as a series of patches and utilizes position embeddings. Chen et al. \cite{chen2021transunet} introduced TransUNet, incorporating ViT as an encoder in the U-Net architecture and significantly improved segmentation performance. 
Due to the absence of displacement invariance and local modeling, training transformers need large-scale data, for example, pure ViT requires millions of images to train. 
Therefore, some recent work has focused on the convolution-transformer fusion structure. Aravind et al.  \cite{srinivas2021bottleneck} propose a simple but powerful instance segmentation and object recognition backbone called BoTNet, which employs MHSA to replace some of the convolutional layers and performs well. 
Nevertheless, these Transformer methods still have the problem of excessive computational cost and GPU memory usage. Therefore, it is necessary to design a low computational and high-efficiency transformer structure without a large amount of data or pre-training weight.


\section{Methods}
In this section, we will give a detailed description of the proposed framework for unsupervised domain adaptation. 
The whole architecture of the proposed framework is shown in \textcolor{ref}{Fig. \ref{fig:network}}, which mainly consists of three modules: dispensed residual transformer block, multi-scale consistency regularization, and feature ranking discriminator. 

\subsection{Architecture Overview}
The essence of our DTNet is to build one deep network that is capable to achieve both multi-domain feature extraction and feature distribution alignment in the latent space. 
As illustrated in \textcolor{ref}{Fig. \ref{fig:network}}, during the source training procedure, inputs are fed into a CNN encoder that disperses representations into latent space.
Next, scattered features are interleaving computed by the dispensed residual transformer block, and finally the global feature operation is realized with small computational cost.
Solid inferences are made considering multi-scale consistency, which alleviates the loss of details in the low-resolution output for better feature alignment. The feature ranking discriminator selects the most valuable features, then discriminates which domain it belongs to. The segmentation backbone is expected to confuse the discriminator and minimize the discriminator's loss by selecting domain invariant features. 


\subsection{Dispensed Residual Transformer Block}
\begin{figure}[ht]
    \centering
    \includegraphics[width=.45\textwidth]{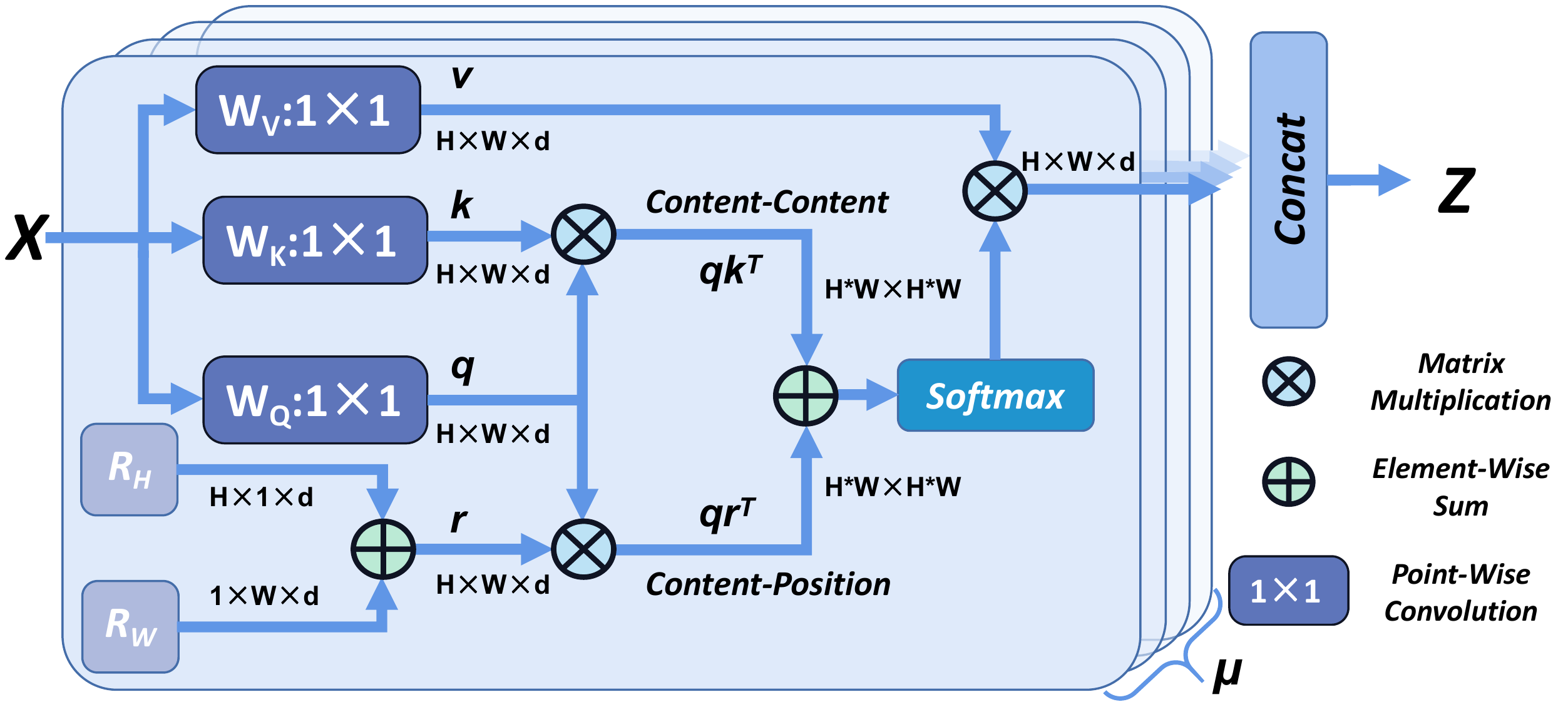}
    \caption{Multi-head self-attention layer structure, where $\mu$ represents the number of heads, and X and Z denote the input and output of the MHSA, respectively.}
    \label{fig:self_att_layer}
\end{figure}

At the bottom layer of the segmentation network, the dispensed residual transformer block is deployed to do non-local feature extraction. Strong representation capabilities of the transformer highly rely on the self-attention mechanism, which is described in Eq. (\ref{f3}).

\begin{equation}
\begin{aligned}
Attention(Q,K,V)=softmax(QK^T/\sqrt{d_k})V 
\end{aligned}
\label{f3}
\end{equation}
where $d_k$ refers to the dimension of the key. It is known as self-attention when $Q$ (queries), $K$ (keys), and $V$ (values) are equal. The detail of our used multi-head self-attention (MHSA)\cite{vaswani2017attention} is shown in \textcolor{ref}{Fig. \ref{fig:self_att_layer}}, in which our position encoding method is relative-distance-aware position encoding~\cite{shaw2018self,ramachandran2019stand,Bello_2019_ICCV} with $R_{h}$ and $R_{w}$ representing height and width respectively. The attention logit is $qk^T$ + $qr^T$, and $q, k, v, r$ denote query, key, value, and position encodings, respectively.

In practice, transformer in vision often wrestles with high computational costs and low training stability. One of the reasons is that the wide and fully connected layers employed by the self-attention create a high dimension sparse feature space that is hardly optimized, weakly generalized, and computationally intensified. Thus, directly utilizing original MHSA is expensive. To address this problem, we present dispensed MHSA, which divides features into groups following three different policies, including channel dispensation, neighbor dispensation, and dilated dispensation.
Our three policies cooperate and complement each other, interleave operations, and finally realize global attention without high computational cost and GPU memory usage of the transformer. In addition, there is a $3 \times 3$ convolution before and after each MHSA constructs the bottleneck structure.

\begin{figure}[ht]
    \centering
    \includegraphics[width=.5\textwidth]{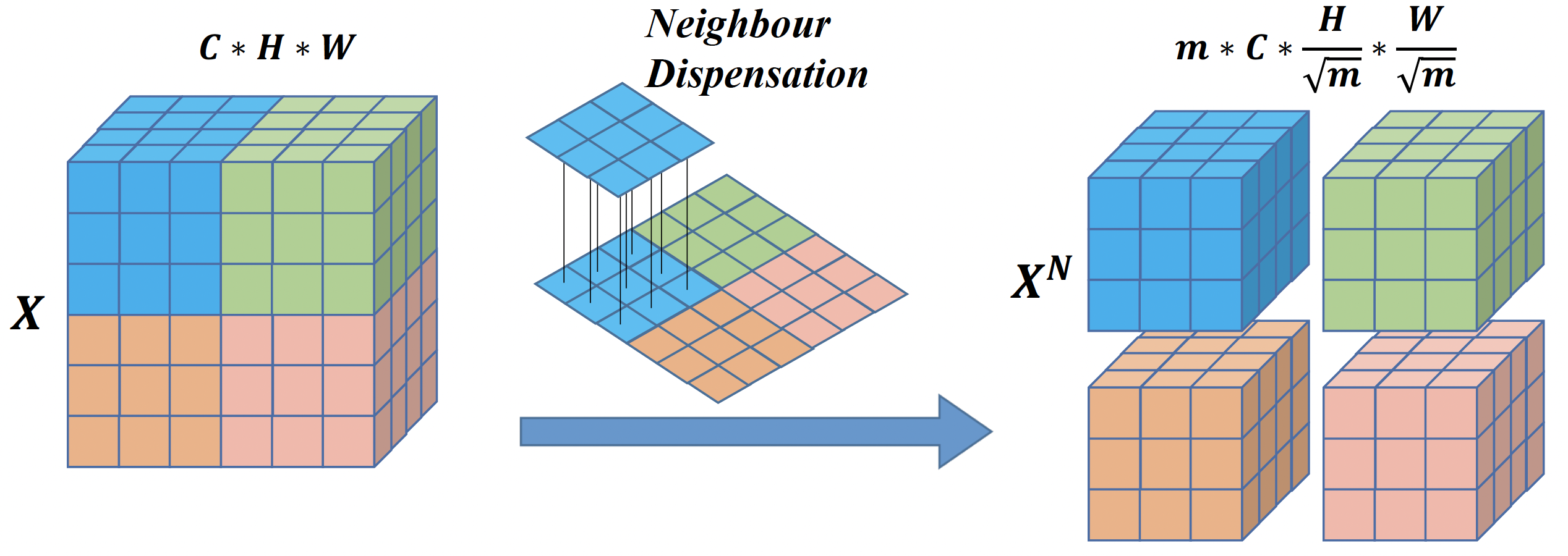}
    \caption{Demonstration of neighbour dispensed multi-head self-attention.}
    \label{fig:NMHSA}
\end{figure}

\subsubsection{Neighbour Dispensation} 
Neighbour dispensed multi-head self-attention (NMHSA) decomposes the feature blocks into subsets containing short spatial interval distances. 
The goal of NMHSA is to calculate the self-attention in the local range and calculate the correlation between each point and its surrounding points, which is very important for the segmentation task.
According to \textcolor{ref}{Fig. \ref{fig:NMHSA}}, the feature map $X$ is firstly divided into $m$ divisions with the same size $\frac{H}{\sqrt{m}}\times \frac{W}{\sqrt{m}}$, described as $X^N = [X^N_1, ..., X^N_i, ..., X^N_m]$, then MHSA is operated on each division $X^N_i$ to generate $A^N$, which is merged to the original size as the output after MHSA.

\begin{figure}[ht]
    \centering
    \includegraphics[width=.5\textwidth]{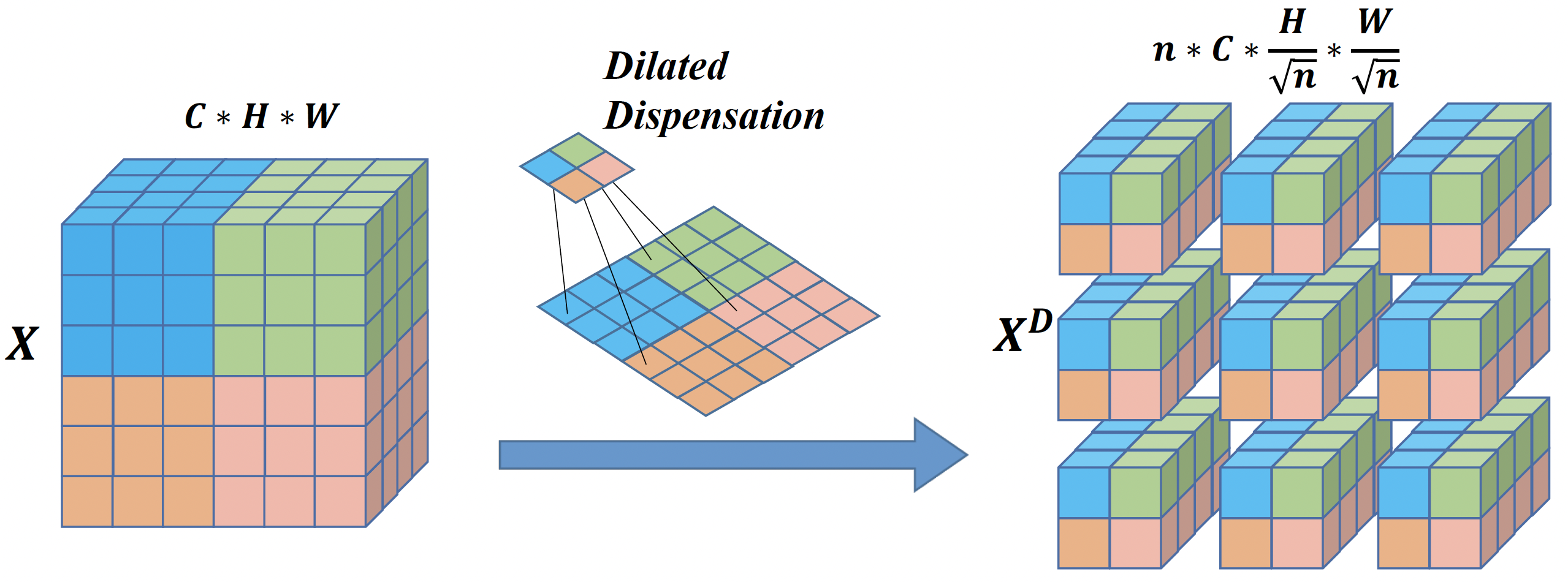}
    \caption{Demonstration of dilated dispensed multi-head self-attention.}
    \label{fig:DMHSA}
\end{figure}


\subsubsection{Dilated Dispensation}
Dilated dispensed multi-head self-attention (DMHSA) tries to squeeze non-local characteristics efficiently from the whole feature block. 
Presented in \textcolor{ref}{Fig. \ref{fig:DMHSA}}, its dilated rate is the width of the neighbor dispensed divisions $\frac{H}{\sqrt{m}}$ or $\frac{W}{\sqrt{m}}$.
The input feature $X$ generates a dilated feature map $X^D$ then regroups it into $n$ blocks with long visual gaps, described as $X^D = [X^D_1, ..., X^D_i, ..., X^D_n]$. 
Each block in $X^D$ is with the size of $\frac{H}{\sqrt{n}}\times \frac{W}{\sqrt{n}}$, containing locations from diverse parts of the whole image. It is worth noting that the multiplication of $m$ and $n$ is equal to $H$ or $W$.
Each block uses MHSA independently to generate $A^D$, which is merged to the original size as the output after MHSA.

\begin{figure}[ht]
    \centering
    \includegraphics[width=.5\textwidth]{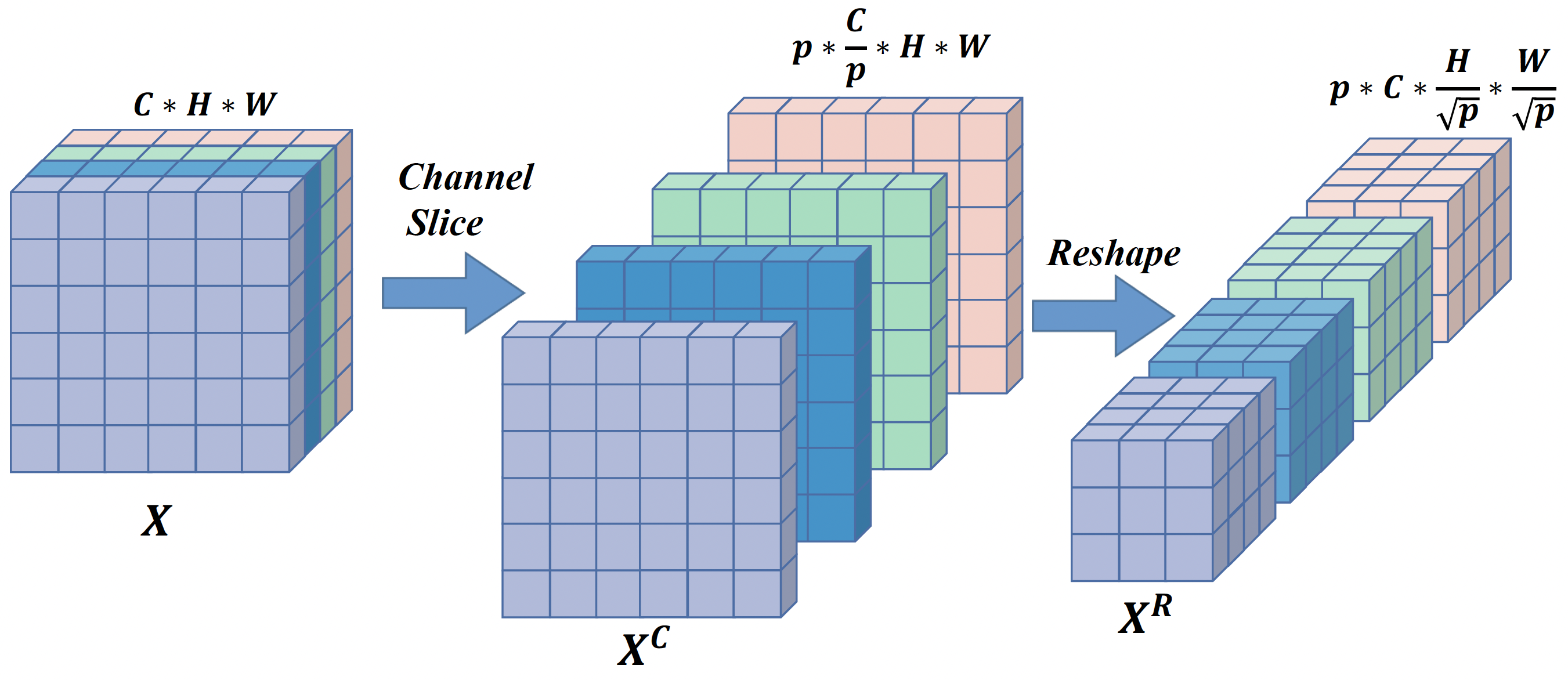}
    \caption{Channel dispensed multi-head self-attention is demonstrated in this image.}
    \label{fig:CMHSA}
\end{figure}

\subsubsection{Channel Dispensation}
Channel dispensed multi-head self-attention (CMHSA) provides the channel-wise features grouping from NMHSA and DMHSA, which devotes to searching valuable features at each channel of the feature block horizontally. This not only provides a new grouping perspective and further improves our global feature extraction, but also the separate operation of different channel features helps to decouple different features, helping to reduce the performance loss of the domain shift.

The main concept of CMHSA is decomposing the feature block into equal-sized blocks according to its number of channels. As presented in \textcolor{ref}{Fig. \ref{fig:CMHSA}}, we slice feature block $X$ from channel to $X^C$, which has $p$ identical sized block of features, described as $X^C = [X^C_1, ..., X^C_i, ... X^C_p]$. Then, we reshape each feature block $X^C_i$ to $X^R_i$ with the size of $\frac{H}{\sqrt{p}}\times \frac{W}{\sqrt{p}}$. 
Although channel slice can reduce the computational cost of MHSA to a certain extent, the GPU memory usage of MHSA is only related to $H$ and $W$ of the feature blocks, independent of the number of channel $C$. Therefore, in order to reduce the memory usage, we also reshape the feature blocks $X^C_i$ after the channel slice.
Multi-head self-attention is operated on $X^R_i$ for a better projection of the global features. After the computation of MHSA, all the feature blocks are reshaped and merged to the original size as the output.

\subsubsection{Calculational Cost} 
The computational cost of the original MHSA before improving \cite{liu2021swin} and that of our NMHSA by the \textcolor{ref}{Fig. \ref{fig:NMHSA}} are given in Eqs. (\ref{eq_MHSA_computational_complexity}) and  (\ref{eq_NMHSA_computational_complexity}) respectively:
\begin{equation}
    \begin{aligned}
        { \Omega  \left( MHSA \left) =4HWC\mathop{{}}\nolimits^{{2}}+2 \left( HW \left) \mathop{{}}\nolimits^{{2}}C\right. \right. \right. \right. }
    \end{aligned}
    \label{eq_MHSA_computational_complexity}
\end{equation}
\begin{equation}
        \begin{aligned}
 \Omega  ( NMHSA ) &=m \times 4 \times \frac{H}{\sqrt{m}}  \times \frac{W}{\sqrt{m}} \times \left( \frac{C}{\lambda} \right) \mathop{{}}\nolimits^{2}\\
{}& + m \times 2 \times  \left( {\frac{H}{{\sqrt{m}}} \times \frac{W}{{\sqrt{m}}}} \right) \mathop{}\nolimits^{2} \times \frac{C}{\lambda}  \\
&=\frac{4}{ \lambda \mathop{}\nolimits^{2}}HWC\mathop{}\nolimits^{2}+\frac{2}{m \lambda }{ \left( {HW} \right) }\mathop{}\nolimits^{2}C
        \end{aligned}
    \label{eq_NMHSA_computational_complexity}
\end{equation}
where $\lambda$ is the scaling factor of the bottleneck structure and $m$ is the neighbour dispensation factor. Same to NMHSA, CMHSA and DMHSA can be simply given in Eqs. (\ref{eq_CMHSA_computational_complexity}) and  (\ref{eq_DMHSA_computational_complexity}):

\begin{equation}
    \begin{aligned}
        { \Omega  \left( DMHSA \right) =\frac{{4}}{{ \lambda \mathop{{}}\nolimits^{{2}}}}HWC\mathop{{}}\nolimits^{{2}}+\frac{{2}}{{n \lambda }}{ \left( {HW} \right) }\mathop{{}}\nolimits^{{2}}C }
    \end{aligned}
    \label{eq_DMHSA_computational_complexity}
\end{equation}

\begin{equation}
    \begin{aligned}
        { \Omega  \left( CMHSA \right) =\frac{{4}}{{ \lambda \mathop{{}}\nolimits^{{2}}}}HWC\mathop{{}}\nolimits^{{2}}+\frac{{2}}{{p \lambda }}{ \left( {HW} \right) }\mathop{{}}\nolimits^{{2}}C }
    \end{aligned}
    \label{eq_CMHSA_computational_complexity}
\end{equation}
in which $n$ and $p$ are the dilated dispensation factor and the channel slice reshape factor, respectively. 
Considering that our network follows a U-shaped structure, $H$ and $W$ of the input feature block of CMHSA are $\frac{1}{2} $ of the other two, and the number of the channels is twice that of them.
For MHSA, the computational cost of the original transformer and that of our dispensed residual transformer are computed through Eqs. (\ref{eq_original_MHSA_total_complexity}) and  (\ref{eq_DRT_computational_complexity}): 

\begin{equation}
    \begin{aligned}
        { \Omega \left( T \right) = 12HWC\mathop{{}}\nolimits^{{2}}+{ \left( {2+\frac{{1}}{{4}}+2} \right) }{ \left( {HW} \right) }\mathop{{}}\nolimits^{{2}}C }
    \end{aligned}
    \label{eq_original_MHSA_total_complexity}
\end{equation}

\begin{equation}
    \begin{aligned}
    \Omega  \left( DRT \right) &= \Omega  \left( NMHSA \right) + \Omega  \left( CMHSA \right)+ \Omega  \left( DMHSA \right)\\
    &=\frac{{12}}{{ \lambda \mathop{{}}\nolimits^{{2}}}}HWC\mathop{{}}\nolimits^{{2}}+{ \left( {\frac{{2}}{{\lambda m}}+\frac{{1}}{4\lambda p}}+\frac{2}{\lambda n} \right) }{ \left( {HW} \right) }\mathop{}\nolimits^{2}C
    \end{aligned}
    \label{eq_DRT_computational_complexity}
\end{equation}
In order to more clearly compare the reduction of the computational cost, let $g_1 = max(\lambda, m, p, n)$, and $g_2= min(\lambda, m, p, n)$. From the above equation, we have:
\begin{equation}
    \begin{aligned}
    \frac{{1}}{{ g_1 \mathop{{}}\nolimits^{{2}}}}\Omega  \left( T \right)\leq
        \Omega  \left( DRT \right) \leq
        \frac{{1}}{{ g_2 \mathop{{}}\nolimits^{{2}}}}\Omega  \left( T \right)
    \end{aligned}
    \label{approxeq_original_MHSA_total_complexity}
\end{equation}

\subsubsection{GPU Memory Usage}
The other main obstacle of implementing the transformer in computer vision tasks lies in significant GPU memory usage. As shown in \textcolor{ref}{Fig. \ref{fig:self_att_layer}}, the original MHSA leads to a high memory occupation largely due to performing content-content and content-position calculations, to produce the matrix, attention map $qk^T$ and $qr^T$ with the size of $ H\times W \times H\times W$. It is clear that NMHSA and DMHSA can significantly reduce its memory usage by reducing $H$ and $W$.
Different from NMHSA and DMHSA, pure channel slice only decreases channel $C$ but not $H$ and $W$, resulting in memory increase by $p$ times, thus the operation of reshaping before CMHSA is essential. As shown in \textcolor{ref}{Fig. \ref{fig:CMHSA}}, we reshape the input feature map of MHSA to ${\frac{H}{\sqrt{p}} \times \frac{W}{\sqrt{p}}}$, which reduces the memory consumption as much as possible on the premise of realizing the operation of multi-head self-attention of the feature blocks. 


\subsection{Multiscale Consistency Regularization}
Visual frequency in different spatial resolutions has a notable variance. Predictions made in low resolutions usually concentrate on semantic level characteristics and identifying semantic level features is a dominant task in higher-level features. Following this on, it is possible for a model to improve its segmentation performance by making a correct prediction on various scales of feature blocks that have a diverse spatial frequency.
As shown in \textcolor{ref}{Fig. \ref{fig:network}}, when training on the source domain, we produce multi-scale segmentation $p_1, p_2, ... p_t$ by picking up different sizes of feature blocks (i.e. $f_1, f_2, ..., f_t$), upsample $f_1, f_2, ..., f_{t-1}$ to the same size of $f_t$ and then add a prediction layer implemented by $1\times 1$ convolution. The $i_{th}$ $(1\leq i \leq t-1)$ uncertainty is quantified by KL-divergence between the top scale prediction $p_t$ and predictions at other scale $p_i$:
\begin{equation}
    {U}_{i} \approx  p_{i} \log \frac{p_{i}}{p_{t}}
\end{equation}
where $U_i$ is the corresponding uncertainty map approximated between $p_i$ and $p_t$. The total number of uncertainty maps is $t-1$, and multiscale consistency loss $L_{mc}$ is formulated as:
\begin{equation}
    L_{\text {mc}}=\frac{1}{t-1} \left( \sum_{i=1}^{t-1}\left\|U_{i}\right\|_2 + \frac{\sum_{i=1}^{t-1} \left(p_{i}-p_{t}\right)^{2} \cdot e^{-U_i} }{\sum_{i=1}^{t-1} e^{-U_i} } \right)
\end{equation}
Following the idea of \cite{zheng2021rectifying,luo2021efficient}, the first term is used to minimize the overall uncertainty and produce more reliable predictions. The second term of $L_{mc}$ to enable uncertainty rectification, which employs $e^{-U_i}$ to automatically give a higher weight to low uncertainty pixels and lower weight for high uncertainty ones.

\subsection{Feature Ranking Discriminator}
According to \cite{ben-david2010a}, if the source and target domains are generalizable, the performance shift of the classifier operated on both the domains can be reduced by shrinking the data distribution distance. As a result, it is effective to employ an adversarial objective (i.e. domain discriminator) to confuse the features learned from different domains. Our approach extends this concept by proposing a ranking module for giving more weights to the more informative feature then generating a discriminator loss $L_{dis}$ to guide the target domain training.

In the ranking branch, we firstly upsample feature blocks $f_1, f_2, ..., f_{t-1}$ to those of the same size and channel as $f_t$ and concatenate them as $F$ with $t*\beta$ channels. 
Feature blocks in $F$ are downsized by global average pooling then two $1\times 1$ convolutions are deployed to generate a path-wised ranking vector $R$. The production of $R$ can be represented as follows:
\begin{equation}
    R=\sigma\left(w_{2} \gamma\left(w_{1} F\right)\right)
\end{equation}
in which $\sigma$ and $\gamma$ are sigmoid and ReLU activations, $w1$ and $w2$ are the convolutional layers used in the ranking branch. According to the score of each feature map, they are ranked in a decreasing order.
The feature blocks from high to low are sent to the corresponding convolutions $C1$, $C2$ and $C3$, where the number of the input channels is identical, but the number of the output channels is $\varphi$, $2\varphi$ and $3\varphi$ respectively. Then, the feature block output by the three paths is connected into the feature block of $6\varphi$ and sent to the domain discriminator, which is a set of convolutions producing a probability map $P_{dis}$ to be compared with domain labels $Y_d$. 

The discriminator loss $L_{dis}$ can be described as
\begin{equation}
    L_{dis} = -(Y_{dis} \log (P_{dis}) + (1-Y_{dis}) \log (1-P_{dis}))
\end{equation}
where $Y_{dis}$ represents the domain label and $Y_{dis}$ is a binary label which has the same size as $P_{dis}$.

\subsection{Overall Loss Function}
In order to fulfill the aim of the unsupervised domain adaptation, our source and target domain training have various loss settings. In the source domain training phase, segmentation loss $L_{ce}$ and multiscale consistency loss $L_{mc}$ are calculated, described as:
\begin{equation}
    L_{source} = L_{ce} + \delta L_{mc}
\end{equation}
where $L_{ce}$ is the cross-entropy loss between the inferred segmentation and the ground truth, $\delta$ is scale constant that controls the ratio of different losses.   When training on the target domain, we use $L_{dis}$ as the target loss to guide the model for reducing the domain shift, and the gradient of the discriminator is set to false, shown as:
\begin{equation}
    L_{target} = L_{dis}
\end{equation}
During the training of the domain discriminator, the gradient of the discriminator is set to true, and the discriminator is trained to classify whether the image is from the source or target domain.

\begin{table*}[ht]
\centering
\renewcommand\arraystretch{1.2}
\caption{Quantitative Comparison results between Our Method and the other methods for The Retinal Nonperfusion Area Segmentation.}
\setlength{\tabcolsep}{2.6mm}
\begin{tabular}{c|c|cc|cc}
\Xhline{1pt}
Method                  & Adaptation      & IoU(\%) & p-value    & Dice(\%)   & p-value  \\ \hline
No-Adap/U-Net\cite{ronneberger2015u}      & Train on Source       & 46.64±26.51      &  4.5E-3* & 58.35±29.58 & 4.0E-3*  \\
No-Adap/TransUNet\cite{chen2021transunet} & Train on Source & 42.89±25.77 & 2.7E-3*  & 55.26±26.88 & 1.4E-3* \\
No-Adap/\textbf{DTNet (ours)} & Train on Source & \textbf{50.89±21.13}  & - & \textbf{64.31±22.86} & -  \\ \hline
CycleGAN\cite{zhu2017unpaired}  & Implicit Feature Alignment    & 32.67±23.54  & 4.0E-9* & 44.58±27.32 & 8.7E-10*  \\
CFDnet\cite{wu2020cf}  & Explicit Domain Discrepancy    & 43.76±19.59 & 1.2E-4*& 58.00±21.83 & 1.2E-7*  \\
PSIGAN\cite{jiang2020psigan}              & Implicit Feature Alignment   & 54.66±21.40  & 7.4E-3*  & 67.69±22.13& 4.5E-2* \\
\textbf{DTNet (ours)}& Implicit Feature Alignment   & \textbf{55.84±18.65}    & - & \textbf{69.53±18.71} & -  \\ \hline
Target Model/U-Net & Train on Target    & 54.16±23.79 &  9.0E-3* & 66.47±25.02 & 9.8E-3*  \\
Target Model/TransNet & Train on Target  & 51.14±22.11  & 9.8E-3* & 64.32±23.09 & 4.7E-2*  \\
Target Model/\textbf{DTNet (ours)} & Train on Target& \textbf{57.33±20.82} & - & \textbf{70.01±22.04} & - \\ \hline
\Xhline{1pt}
\end{tabular}
\label{tab_sota_comparison}
\end{table*}

\begin{figure*}[htbp]
\centering
    \includegraphics[width=0.95\textwidth]{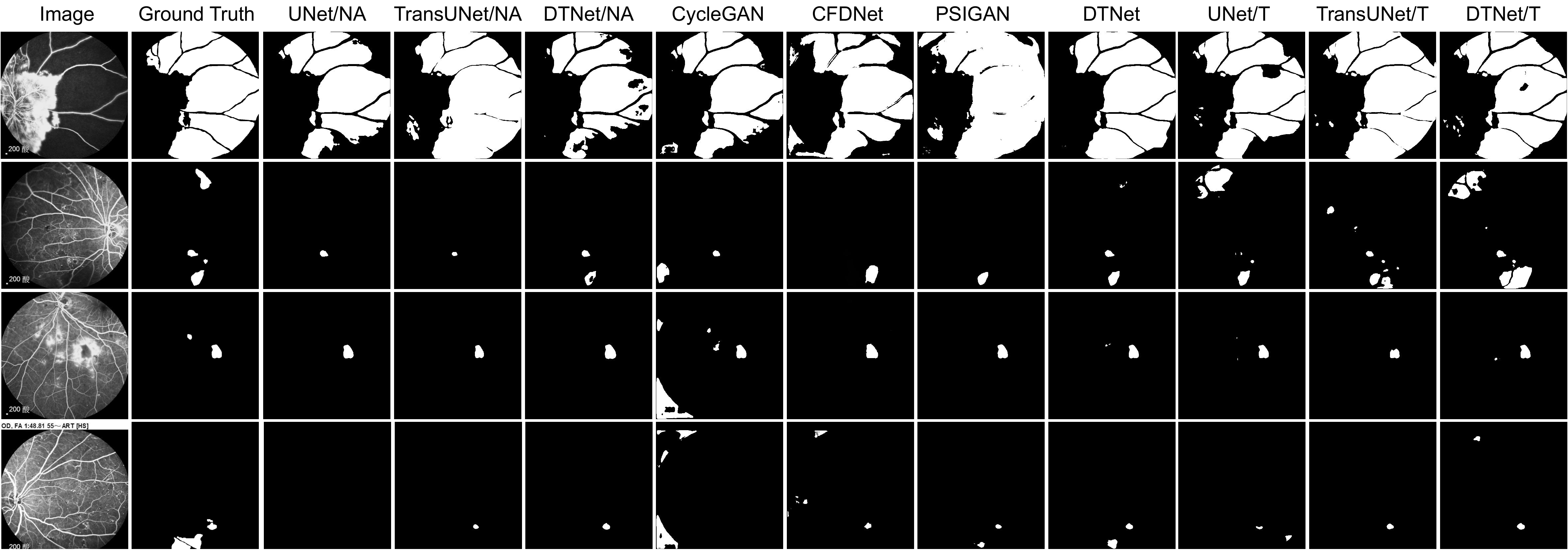}
    \caption{Example of the segmentation results achieved by No Adaptation, Domain Adaptation and Target-trained model for retinal nonperfusion area segmentation.}
    \label{fig_qualitative_evaluations}
\end{figure*}

\section{Experiments and Discussions}

\subsection{Dataset and pre-processing}
\subsubsection{Cross-site Dataset}
We test the performance on both cross-site and the cross-modality datasets, in which the cross-site dataset contains retinal nonperfusion fluorescein angiography images from the Second Affiliated Hospital of Zhejiang University (ZJU) and the Second Affiliated Hospital of Xi'an Jiaotong University (XJTU). ZJU dataset contains 532 images as the source domain, and 144 images from XJTU as the target domain for two-fold cross-validation, all of which are pixel-level annotated by ophthalmologists. All images are resized to $384 \times 384$ with the same size. We perform image transformations for both source and target domain data in all the experiments as augmentations.
Grid distortion, elastic deformation, horizontal flipping, and vertical flipping are randomly used for each case. The probability for doing rotation, horizontal and vertical reflection is 0.5 and for the others is 0.2. 
Note that the data augmentation we applied is only in the training phase while in the validation and testing phase we only resize and normalize the images.

\subsubsection{Cross-modality Dataset}
We verified our method on MM-WHS cross-modality  dataset\cite{zhuang2019evaluation}, which contains 20 CT and MR images with manually segmented ground truth. The test set contains 32 CT and 25 MR cases without annotations. For a fair comparison, we used the official code of CFDnet \cite{wu2020cf} to ensure the same experimental settings and data division. Following the CFDnet \cite{wu2020cf}, we performed two-fold cross-validation by randomly selecting 10 annotated MR images as test data. All the remaining MR images and CT images are used to train our model. For each 3D image, we sampled 16 slices from the long-axis view around the center of the left ventricular cavity. These slices were then cropped with the size of $192\times 192$ pixel around the center of the heart.

\subsection{Implementation Details}

We trained our model via standard Adam optimizer, with an initial learning rate of $10^{-3}$ for a total of 200 epochs for training. All networks were implemented using the Pytorch library and we ran the experiments on a machine equipped with a 32-Core Intel(R) Xeon(R) Silver 4110 CPU, 128 GB of RAM, and 4 NVIDIA RTX 2080Ti GPUs.
Our code is available at \href{https://github.com/Kent0n-Li/DTNet}{https://github.com/Kent0n-Li/DTNet}.

\subsection{Comparision Metrics}
Segmentation accuracies were measured using IoU, Dice, and ASSD, and they are computed via the following:

\begin{equation}
    \begin{aligned}
        IoU = \frac{TP}{FN+FP+TP}
    \end{aligned}
    \label{eq_js}
\end{equation}

\begin{equation}
    \begin{aligned}
        Dice = \frac{2 \times TP}{FP+2 \times TP+FN}
    \end{aligned}
    \label{eq_dc}
\end{equation}

\begin{equation}
    \begin{aligned}
        ASSD = \frac{ \sum_{v_P\in S(P) } d(v_P, S(G))+\sum_{v_G\in S(G)}d(v_G,S(P))}{|S(P)|+|S(G)|} 
    \end{aligned}
    \label{eq_dc}
\end{equation}
where TP, FP, TN, FN are true positives, false positives, true negatives and false negatives. 
Meanwhile, $\ d(v_P, S(G))$ is the shortest distance of an arbitrary voxel $v_P$ to $S(G)$, which is defined as $d(v_P,S(G)) = min_{v_G \in {S(G)}} \|v_P-v_G\|$, and $S(P)$ represents the surface voxels set of $P$.
Considering that retinal nonperfusion segmentation and cardiac segmentation are different, we assign appropriate comparison metrics to examine them.

\subsection{Performance on Cross-site Dataset}

\subsubsection{Lower and Upper Baselines}
We trained segmentation networks in a fully supervised manner in both source and target domains to assess the impact of domain shift. UNet \cite{ronneberger2015u}, TransUNet \cite{chen2021transunet}, and our DTNet without DA trained on the source or target domain serves as lower or upper baseline respectively. With the unsupervised domain adaptation settings, we report four results including CycleGAN \cite{zhu2017unpaired}, CFDnet \cite{wu2020cf}, PSIGAN \cite{jiang2020psigan}, and our DTNet.
We evaluate a CycleGAN-based method, using CycleGAN for domain alignment. First, we train an UNet with source images and report the testing metrics on the domain-alignment target images processed by CycleGAN. We refer this method to as CycleGAN.

\subsubsection{Performance Comparison}
\textcolor{ref}{Table \ref{tab_sota_comparison}} reports the quantitative segmentation results of the retinal nonperfusion area cross-sites dataset.
Further, we performed a one-tail paired t-test between our method and other methods in each part. All the comparing methods whose difference with our method is statistically significant (p-value$\textless 0.05$) are marked by the symbol *.
Our DTNet outperforms all the comparison methods in terms of IoU and Dice.  We also experiment with the methods that are only trained on the source data without adaptation, called No-Adap.
We can see that the No-Adap results are poor, indicating a large domain gap between the different domain images. Due to domain shift, a severe performance drop was observed when migrating model training from the source domain to the target domain. 
It is worth noting that TransUNet is undesirable, and the possible reason is that TransUNet adopts ViT as the encoder. However, ViT relies on a huge image corpus, leading to undesirable performance on insufficient datasets.
We can see that all the networks with domain adaptation have improved a lot, but our DTNet obtains better accuracy.
Among them, CFDnet performs poorly because CFDnet reduces the distribution discrepancy by a metric based on characteristic functions of distributions, limiting their broad applicability. Consequently, CFDnet cannot deal with retinal nonperfusion area segmentation without a fixed similar shape. As shown in \textcolor{ref}{Fig. \ref{fig_qualitative_evaluations}}, the retinal nonperfusion areas are mostly very small, which needs a global attention structure to correctly locate the lesion area, and that is why our DTNet backbone achieved better performance.
Moreover, our model is the closest to the upper baselines trained directly in the target domain, which means that our model achieves the best domain adaptation outcome and reduces the domain shift as much as possible.

\begin{table}[ht]
\renewcommand\arraystretch{1.1}
\caption{Performance Comparisons for The CT-MR Cross-modality Cardiac Segmentation.}
\centering
\label{CFDnet}
\begin{tabular}{c|cc|cc}
\Xhline{1pt}
\multirow{2}{*}{Methods} & \multicolumn{2}{c|}{LV} & \multicolumn{2}{c}{MYO} \\ \cline{2-5}  
 &Dice(\%)  & ASSD(mm)   & Dice(\%)  & ASSD(mm)   \\ \hline
 NoAdapt       & 44.4±13.9  & 19.1±6.52 & 24.4±9.11   & 17.3±2.70 \\
 PnP-AdaNet  & 86.2±6.46 & \textbf{2.74±1.04}  & 57.9±8.43 & \textbf{2.46±0.661} \\
 AdvLearnNet & 83.8±10.3 & 5.76±6.07  & 61.9±15.2 & 3.79±2.23  \\
 MMDnet      & 86.7±8.65 & 3.64±3.37  & 64.4±12.0 & 3.85±2.39  \\
 CFDnet      & 88.7±10.6 & 2.99±2.79  & 67.9±8.62 & 3.40±2.75  \\
 \textbf{DTNet (ours)} & \textbf{89.2±5.67}  &2.84±1.74 & \textbf{70.0±9.59} & 2.56±1.26\\ \Xhline{1pt}
\end{tabular}
\end{table}


\subsection{Performance on Cross-modality Dataset}
\begin{figure}[ht]
    \includegraphics[width=0.5\textwidth]{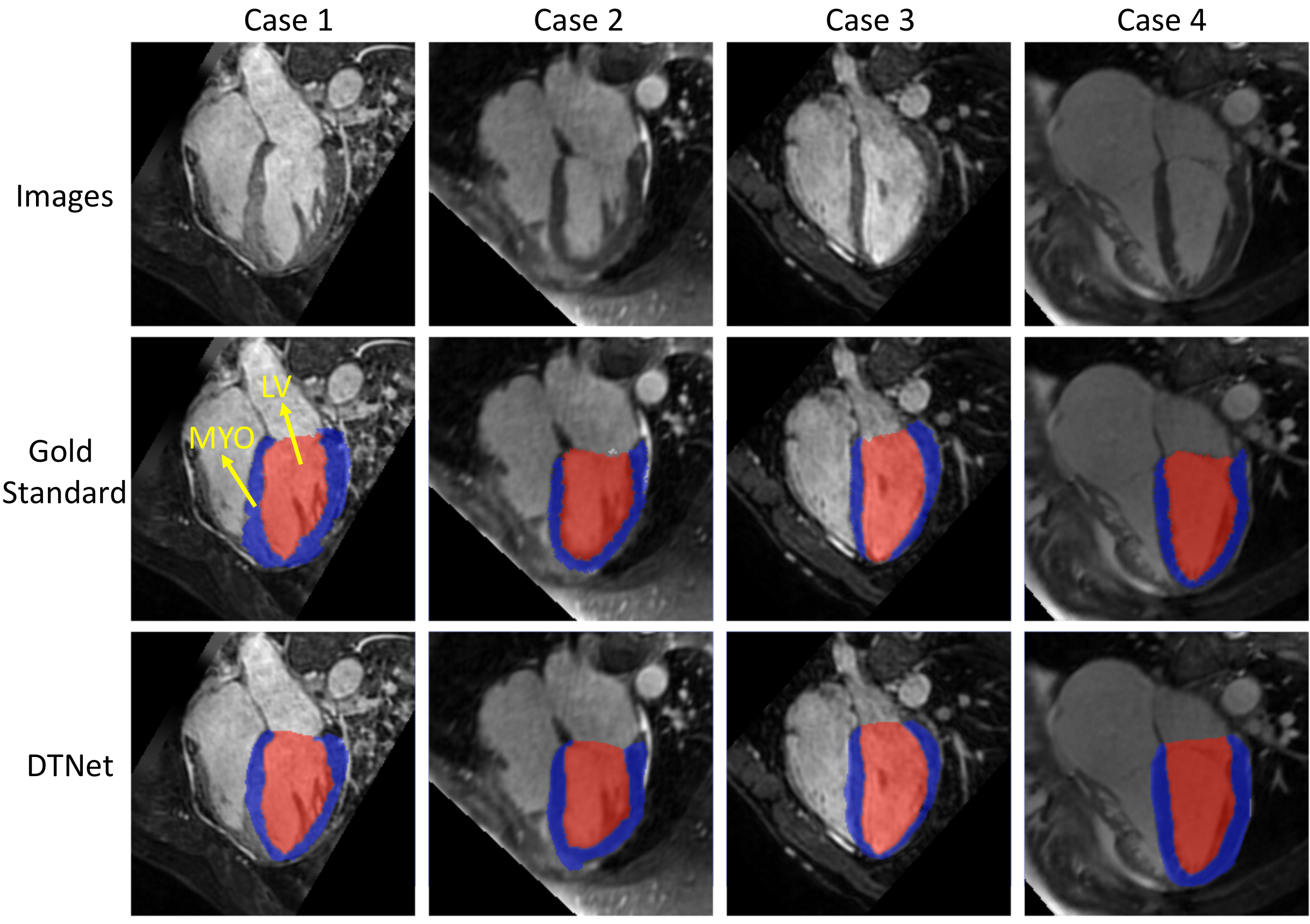}
    \caption{Segmentation results of CT-MR cross-modality cardiac segmentation.}
    \label{mrseg}
\end{figure}

An adaptation model which is suitable for multi-tasks shows more robustness and generalization in practice. Therefore, we test the performance on the widely used public cross-modality dataset from MM-WHS \cite{zhuang2019evaluation}. Besides, we use the data loading code of CFDnet \cite{wu2020cf} with the same data and experimental settings, so we directly quote its experimental results. The description of the comparative experiment in the CFDnet paper will not be recounted here. We tested the performance of domain adaptation for cardiac MR image segmentation using CT images as source data. Considering that the original results are not available in the CFDnet publication, we did not perform a t-test. As shown in \textcolor{ref}{Table \ref{CFDnet}}, NoAdapt in this task performed poorly, indicating that the domain shift between different modality data is evident. Compared to CFDnet, DTNet obtained much better accuracy on all the structures in all the metrics. In addition, our method achieves the highest Dice and the second-highest ASSD among all the methods, which represents that our method has achieved state-of-the-art performance. \textcolor{ref}{Fig. \ref{mrseg}} shows the segmentation results of randomly selected test images, illustrating that our DTNet can achieve excellent domain adaptation.

\begin{table}[ht]
\renewcommand\arraystretch{1.1}
\caption{Ablation Study}
\setlength{\tabcolsep}{1.1mm}
\centering
\label{CFDnet}
\begin{tabular}{c|cc|cc}
\Xhline{1pt}
\multirow{2}{*}{Methods} & \multicolumn{2}{c|}{LV} & \multicolumn{2}{c}{MYO} \\ \cline{2-5}  
 &Dice(\%)  & ASSD(mm)   & Dice(\%)  & ASSD(mm)   \\ \hline
DTNet w/o Disp    & 83.1±16.6  & 3.71±3.59 & 61.3±18.1   & 3.42±3.91 \\
DTNet w/o Multi & 87.1±7.84 & 3.30±2.12  & 69.4±11.5 & 2.92±1.57 \\
DTNet w/o Rank & 64.9±29.9 & 12.1±19.1  & 44.5±21.4 & 9.73±12.9  \\
 \textbf{DTNet (ours)} & \textbf{89.2±5.67}  & \textbf{2.84±1.74} & \textbf{70.0±9.59} & \textbf{2.56±1.26} \\ \Xhline{1pt}
\end{tabular}
\label{tab4}
\end{table}

\subsection{Ablation Study}
The proposed DTNet utilizes three major techniques to improve the system performance, including dispensed residual transformer block, multi-scale consistency regulation, and feature ranking discriminator. To validate their effectiveness, we perform three ablation studies using the MM-WHS cross-modality dataset \cite{zhuang2019evaluation}. The ablation experimental results are shown in \textcolor{ref}{Table \ref{tab4}}. Specifically, we first study the effect of the dispensed residual transformer block. Then, we demonstrate the effect of multi-scale consistency regulation. At last, we study the effect of the feature ranking discriminator, and the details are elaborated in the following sub-sections.

\subsubsection{Efficacy of Dispensed Residual Transformer Block} 
The dispensed residual transformer models the long-range dependencies in the feature maps as complementary information, and it can force the network to focus its attention on the global regions. To validate its effectiveness for domain adaptation and segmentation, we implement the variant of DTNet without this technique, referred to as DTNet w/o Disp. As \textcolor{ref}{Table \ref{tab4}} shows, DTNet compares favorably to DTNet w/o Disp in terms of the segmentation accuracy of both Dice and ASSD.


\subsubsection{Efficacy of Multi-scale Consistency Regularization} 
Making reliable feature blocks in different scales can improve its segmentation performance as solid inferences are made considering multi-scale consistency, which alleviates the loss of details in the low-resolution output for better feature alignment.
In order to verify its effectiveness in improving performance, we compare the performance of the variant of DTNet without this technique, referred to as DTNet w/o Multi. The results are reported in \textcolor{ref}{Table \ref{tab4}}. We can see that multi-scale consistency can effectively enhance the segmentation performance.


\subsubsection{Efficacy of Feature Ranking Discriminator} 
Instead of putting all the features into the discriminator in the same proportion, a feature ranking discriminator gives more weights to the more informative features. As the feature alignment is crucial to the domain adaptation, we argue that the feature ranking discriminator can improve the potential of DTNet for domain adaptation and segmentation. 
To validate this, we evaluate the performance of DTNet without feature ranking discriminator, denoted as DTNet w/o Rank. As \textcolor{ref}{Table \ref{tab4}} shows, without the feature ranking discriminator, the performance of DTNet degraded significantly.


\begin{table}[ht]
\centering
\renewcommand\arraystretch{1.2}
\setlength{\tabcolsep}{0.3mm}
\caption{Comparison of Memory Usage and FLOPs}
\begin{tabular}{c|cc|cc|cc}
\Xhline{1pt}
 & \multicolumn{2}{c|}{No Dispensed} & \multicolumn{2}{c|}{Dispensed} & \multicolumn{2}{c}{Slice w/o Reshape} \\ \hline
size    & Mem(GB)   & FLOPs(G)  & Mem(GB)     & FLOPs(G)     & Mem(GB)     & FLOPs(G)    \\ \hline
8       & 1.4053     & 0.0002     & 1.4053       & 0.0002        & 1.4053       & 0.0002       \\
24      & 1.4443     & 0.0119     & 1.4092       & 0.0019        & 1.5537       & 0.0116       \\
40      & 1.6592     & 0.0872     & 1.4268       & 0.0053        & 2.3701       & 0.0863       \\
56      & 2.3174     & 0.3282     & 1.4307       & 0.0103        & 4.9990       & 0.3264       \\
72      & 3.8584     & 0.8877     & 1.4521       & 0.0171        & 11.1650      & 0.8847       \\
88      & 6.8428     & 1.9683     & 1.4756       & 0.0255        & 23.1162      & 1.9639       \\
96      & 9.0908     & 2.7814     & 1.4795       & 0.0304        & 32.1084      & 2.7761       \\
104     & 11.9854    & 3.8240     & 1.4990       & 0.0356        & 43.6670      & 3.8177       \\
112     & 15.6221    & 5.1341     & 1.5166       & 0.0413        &  -            &   -           \\
120     & 20.1279    & 6.7572     & 1.5264       & 0.0475        & -             &   -           \\
128     & 25.6104    & 8.7363     & 1.5264       & 0.0540        &   -           &  -            \\
136     & 32.2510    & 11.1246    & 1.5420       & 0.0610        &  -            &   -           \\
144     & 40.1436    & 13.9663    & 1.5732       & 0.0684        &   -           &  -            \\
\Xhline{1pt}
\end{tabular}
\label{tabflops}
\end{table}

\subsection{Computation and Memory Analysis}
\begin{figure}[ht]
    \includegraphics[width=0.5\textwidth]{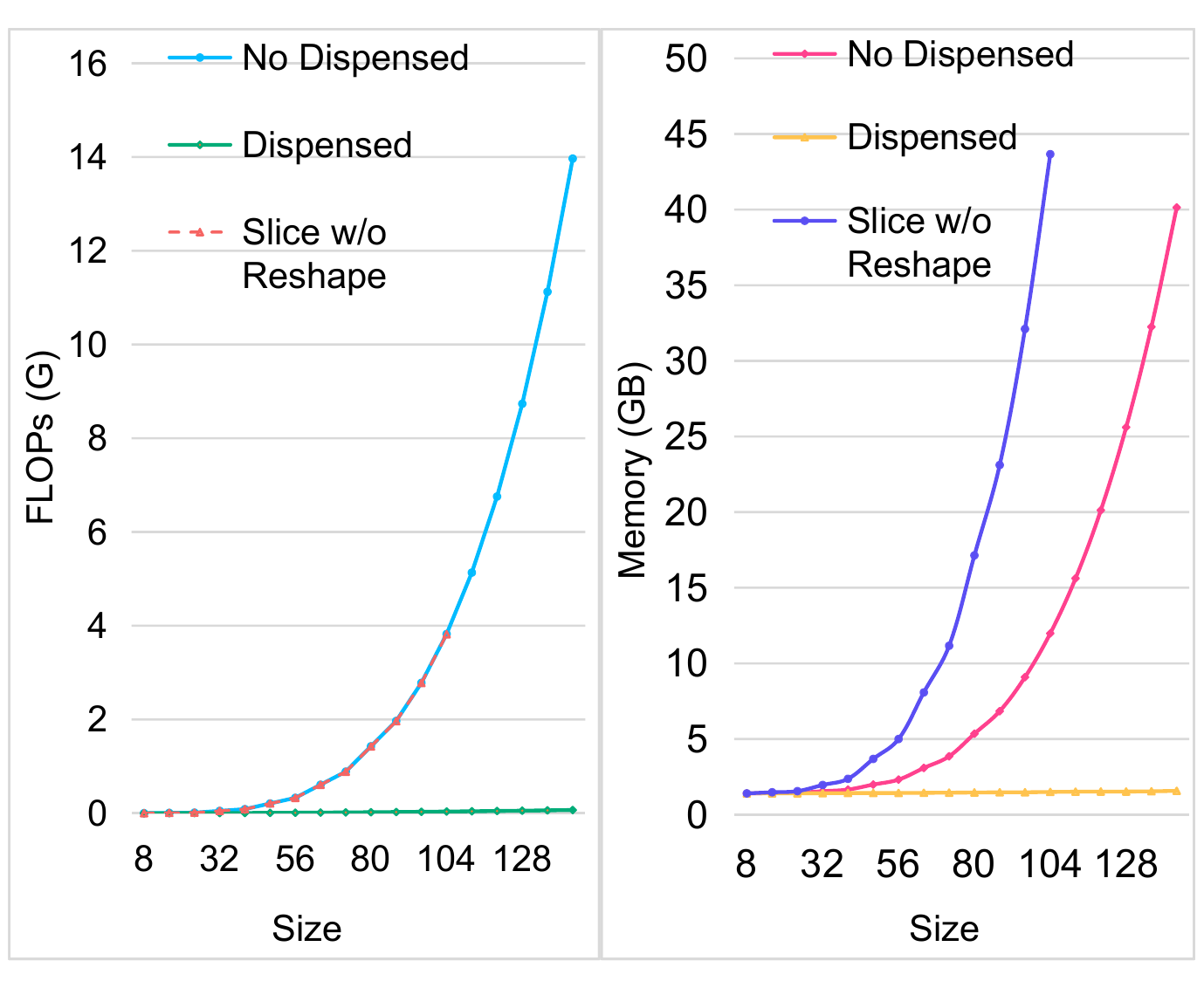}
    \caption{Line chart of the size-flops and size-memory}
    \label{fig_size_flops}
\end{figure}

As we know, the direct application of transformer in computer vision tasks faces the challenge of tremendous computation and memory occupation \cite{matsoukas2021time}. We will compare the effectiveness of our method for two aspects: computational cost and memory occupation. \textcolor{ref}{Table \ref{tabflops}} and \textcolor{ref}{Fig. \ref{fig_size_flops}} depict the changes of memory and FLOPs with the size of input feature blocks from MHSA. We fix the channel as 16 but change the width and height in practice. Columns 'No Dispensed' are original MHSA without dispensing, and columns 'Dispensed' represent our dispensed Transformer. Besides, Columns 'Slice w/o Reshape' represent the channel dispensed without reshaping, but only channels slice.
Note that our three kinds of dispensations have the same complexity-occupation trends of MHSA. 
It is worth noting that only channel slice without reshaping results in memory occupation increasing so sharply that we cannot obtain the GPU memory usage larger than the GPU max memory of our equipment thus denote these blank results as -.

\begin{table}[ht]
    \centering
    \setlength{\tabcolsep}{5.3mm}
    \caption{Comparison on Different Feature Channels for MHSA.}
    \begin{tabular}{c|c|c}
    \Xhline{1pt}
Channel & Mem(GB)  & FLOPs(G) \\ \hline
128    & 1.6729      & 0.77 \\
256    &   1.6982     & 1.69  \\
512        & 1.7236     & 4.01  \\
1024    & 1.7764     & 10.53  \\
2048       & 1.9365      & 31.14  \\
4096      & 2.4170       & 102.54   \\ \Xhline{1pt}
    \end{tabular}
    \label{tab_bottleneck_mem_flops_analysis}
\end{table}


To demonstrate the efficacy of our bottleneck structure, we report the results of the computation and occupation of our dispensed transformer with different feature channel amounts. We set the width and height of the input feature block to 40 when examining the bottleneck structure. As depicted in \textcolor{ref}{Table \ref{tab_bottleneck_mem_flops_analysis}}, the GPU memory usage and FLOPs increase sharply with the increasing channel.

\section{Conclusion}
In this paper, we have presented a novel unsupervised domain adaptation method for the segmentation of cross-site and cross-modality datasets. 
We proposed a dispensed residual transformer to realize global attention by dispensed interleaving operation to address the challenges of the excessive computational cost and GPU memory usage of the transformer. To alleviate the loss of details in a low-resolution output for better feature alignment, a multi-scale consistency regularization was designed.
Also, we introduced a feature ranking discriminator to automatically optimize the weights to the domain-gap features to better reduce the feature distribution distance.
Extensive experiments on cross-site and cross-modality datasets show that our proposed method outperforms the baseline networks, and also achieves better segmentation accuracy than the other state-of-the-art unsupervised domain adaptation methods for retinal nonperfusion area and cardiac segmentation. 

In future work, we will further improve the segmentation performance for the following two aspects. (1) In order to further explore the performance of the transformer, we can remove convolution and construct the whole network with the transformer.
(2) We will optimize the framework and further extend to 3D/4D volume segmentation.

\bibliographystyle{IEEEtran.bst}
\bibliography{paper.bib}

\end{document}